\documentclass[letterpaper, 10 pt, conference]{ieeeconf}  
\usepackage[T1]{fontenc}
\usepackage{amsmath,amssymb,bm}
\usepackage{booktabs}
\usepackage{graphicx}
\usepackage[hidelinks]{hyperref}
\usepackage[font=footnotesize,textfont=normalfont,justification=justified,
            singlelinecheck=false,labelsep=colon]{caption}
\usepackage{xcolor}
\usepackage{dblfloatfix}
\usepackage{microtype}
\microtypesetup{protrusion=false}
\usepackage{tikz}
\usepackage{subcaption}
\usepackage{arydshln}
\usepackage[capitalize]{cleveref}
\usepackage{balance}

\usetikzlibrary{arrows.meta, positioning}
\newcommand\blfootnote[1]{%
  \begingroup
  \renewcommand\thefootnote{}\footnote{#1}%
  \addtocounter{footnote}{-1}%
  \endgroup
}

\title{\LARGE \bf
OTRetarget: Joint Robot and Object Motion Retargeting\\ via Optimal Transport
}
\author{\parbox{\linewidth}{\centering{Guillaume Besset$^{*,1}$, Erwann Carn$^{*,1}$, Timothée Carecchio$^{1}$,
        Valentin Tordjman--Levavasseur$^{1}$, Fabian Schramm$^{1}$, Yann de Mont-Marin$^{1}$, Justin Carpentier$^{1}$, Ajay Suresha Sathya$^{1,2}$}}%
}
\newcommand{\sys}{\textsc{OTRetarget}}

\begin{document}
\makeatletter
\twocolumn[
  \begin{@twocolumnfalse}
    \maketitle
    \begin{center}
        \vspace{0cm}
        \resizebox{1\linewidth}{!}{\includegraphics[trim={19pt 200pt 19pt 220pt}, clip]{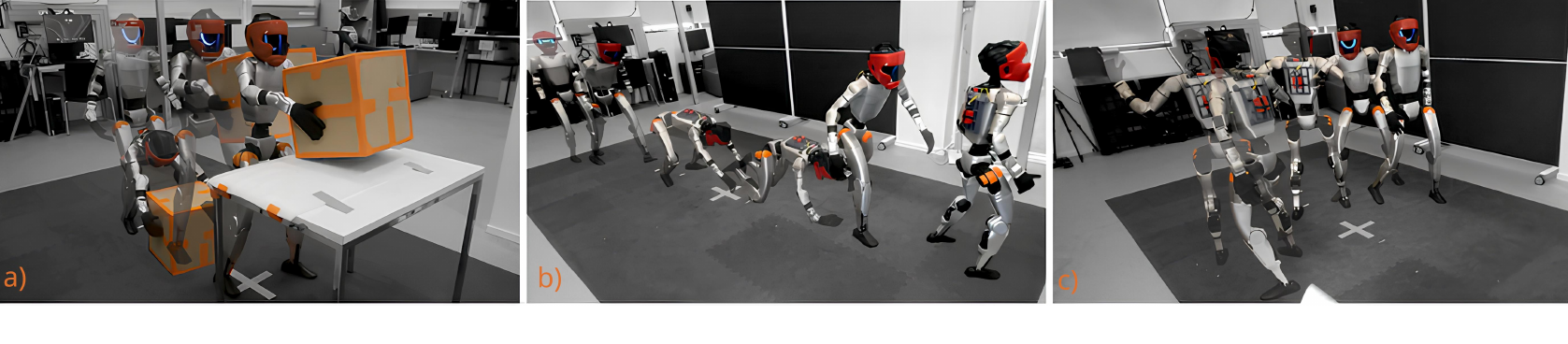}}
        \captionsetup{labelsep=colon,skip=-10pt}
        \captionof{figure}{\textbf{Whole-body loco-manipulation with \sys{}.} Our approach jointly retargets robot and object motion while preserving interactions with the ground and surrounding objects, without rescaling the scene or demonstration. Overlaid poses show successive instants of the retargeted motions: \textbf{a)}~two-handed box pickup and placement onto a table; \textbf{b)} quadrupedal crawling with hand--ground contact; and \textbf{c)} locomotion with a full-body spin.}
        \label{fig:teaser}
    \end{center}
  \end{@twocolumnfalse}
]
\makeatother

\blfootnote{\textsuperscript{*}Equal contribution.}
\blfootnote{\textsuperscript{1}Inria, D\'epartement d'Informatique de l'\'Ecole Normale
  Sup\'erieure, PSL Research University, Paris, France.
  \href{mailto:firstname.lastname@inria.fr}{firstname.lastname@inria.fr}}
\blfootnote{\textsuperscript{2}Dept.\ of Aeronautics and Astronautics, Stanford University,
  CA, USA.}
\blfootnote{Project webpage:
  \href{https://simple-robotics.github.io/publications/otretarget/}{simple-robotics.github.io/publications/otretarget}}

\begin{abstract}
Transferring human motion to humanoid robots requires adapting the demonstrated motion to the robot morphology while preserving interactions with the environment.
This is particularly challenging for loco-manipulation tasks, where contacts with the ground and manipulated objects must remain consistent despite differences in body proportions.
Yet, skeletal motion alone does not fully describe these interactions, and fixing object trajectories limits the adaptation to a new embodiment.
In this paper, we introduce \sys{}, a unified approach to jointly retarget robot and multi-object motion from human demonstrations.
Our approach represents surface interactions through signed distances, closest surface points, and relative directions, and uses entropic optimal transport to transfer these quantities across human, robot, and object geometries.
We incorporate the resulting interaction targets into a constrained inverse kinematics formulation that balances contact preservation with motion style and jointly optimizes robot and object poses at each frame.
This formulation accommodates robot--object and object--object interactions without rescaling the scene or the demonstration.
We validate the proposed approach on OMOMO, where it achieves a robot--object interaction Jaccard score of $87\%$ and a depth error of $8.7$\,mm, compared with $28\%$ and $29.3$\,mm for OmniRetarget.
Finally, we demonstrate transfer to a physical G1 humanoid using whole-body policies trained with reinforcement learning on the retargeted references, across motions including two-handed box pick-and-place onto a table.
\end{abstract}

\section{Introduction}\label{sec:intro}

Imitating human motion has attracted growing interest in character animation and humanoid control \cite{deepmimic2018,phc2023}. Commonly, a human motion is \emph{retargeted} into a kinematic reference trajectory, a \emph{whole-body tracking policy} learns to follow it~\cite{beyondmimic2025}, and the policy is deployed on the physical robot \cite{h2o2024,asap2025,omniretarget2025}.
Because the policy learns from this retargeted reference, its embodiment gap-related artifacts propagate to the learned policy unless reward engineering compensates for them~\cite{gmr2026}.

In free-space locomotion, the embodiment gap is largely one of \emph{scale}: rescaling the source motion, done right, removes most of it \cite{gmr2026}, with reward shaping and domain randomization covering the rest \cite{omniretarget2025}.
Contact-rich motions, such as crawling or carrying a box, make retargeting errors more consequential. If a foot skates or a hand fails to make contact with the payload, learned behavior could fail regardless of joint tracking accuracy.
In whole-body loco-manipulation \cite{omniretarget2025,hdmi2025}, the demonstrated interactions need to be preserved while adapting robot and object trajectories.

This creates three related failure modes.
\mbox{\textbf{(i) Scaling inconsistency.}} Rescaling shrinks the robot and object trajectories, not the scene, so contacts no longer align with the floor or table.
\mbox{\textbf{(ii) Unadapted object trajectory.}} The demonstrated box trajectory lifted from the floor onto a table can be carried at heights unreachable for the robot when unscaled, and might deposit the box \emph{under} the table when heights are rescaled.
\mbox{\textbf{(iii) Interaction loss.}} Even for feasible trajectories, contact must adapt across embodiments: a shorter-armed robot might need to open its arms wider to appropriately grasp a box.

\begin{figure*}[!t]
\vspace{0.5em}

    \centering
    \makebox[\textwidth][c]{\hspace*{-0.8cm}\resizebox{1.055\textwidth}{!}{\input{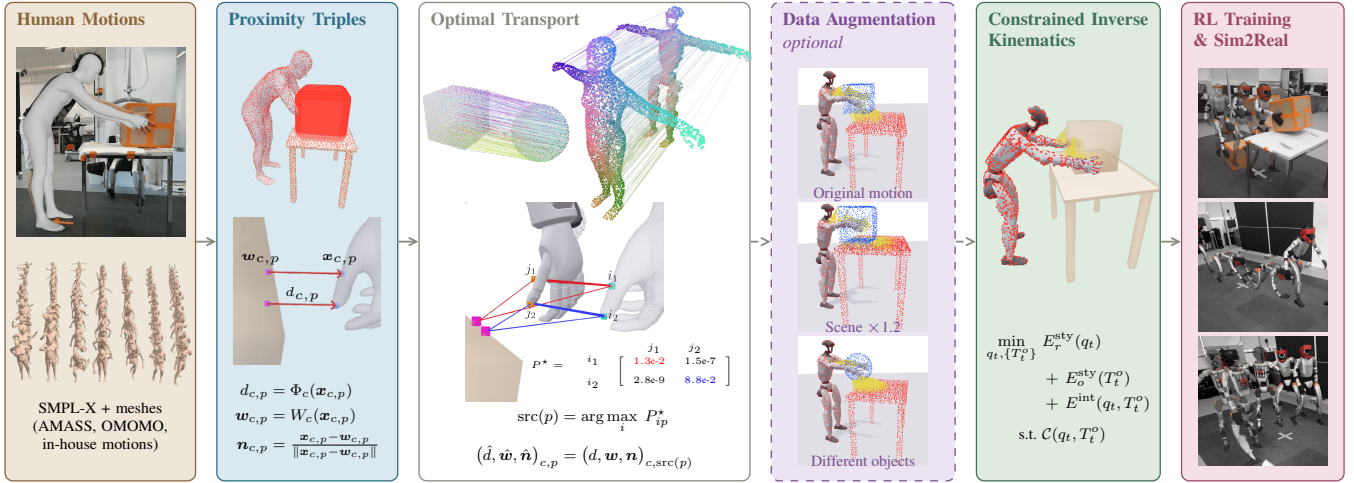}}}
    
    \captionsetup{labelsep=colon}
    \caption{\textbf{Pipeline overview.} Each probe $\bm{x}_{c,p}$ reads a proximity triple $(d,\bm{w},\bm{n})_{c,p}$ from the channel's signed distance field. A transport plan $P^\star$, computed once, transfers this target to the robot. At each frame, a constrained program balances the resulting residuals against the skeleton-style targets while jointly solving the robot pose $q_t$ and every object pose $T^o_t$. Dashed: the optional object substitution (Sec.~\ref{sec:m-datagen}).}
    \label{fig:pipeline}
\vspace{-1em}
\end{figure*}

Existing retargeting methods leave at least one of these challenges unresolved: their reliance on skeletal motion provides only a partial description of the surface interactions that govern contact.
To address these limitations, we introduce \sys{}, a proximity-aware method for whole-body loco-manipulation that jointly adapts robot and object motion using interaction targets extracted from scene geometry (Fig.~\ref{fig:pipeline}).
Our approach operates at the retargeting stage of the imitation pipeline, where human demonstrations are converted into robot reference trajectories.

Our approach represents interactions directly through the geometry of the participating surfaces. Signed distances describe contact and separation, while closest surface points and relative directions capture where and how these interactions occur. We evaluate these quantities at densely sampled surface points and use optimal transport to establish correspondences across human, robot, and object geometries. This representation lets us formulate retargeting as a balance between preserving the demonstrated interactions and retaining the motion style encoded by skeletal and object trajectories.
Our work makes three contributions:

\noindent\textbf{C1. Surface-level interaction residuals.}
We introduce pointwise interaction residuals extracted from demonstration geometry and integrate them with motion-style objectives in a per-frame constrained inverse kinematics problem. Our approach captures surface relationships that skeletal targets alone cannot express, addressing interaction loss across embodiments~(iii).

\noindent\textbf{C2. Joint robot and multi-object retargeting.}
Our formulation jointly optimizes robot and object poses, allowing their relative motion to adapt to the target embodiment while preserving robot--object, object--object, and ground interactions. By treating object trajectories as decision variables, our approach accommodates multiple interacting objects without rescaling the scene or demonstration, addressing scaling inconsistency~(i) and unadapted object trajectories~(ii).

\noindent\textbf{C3. Interaction transfer across object geometries.}
We extend the optimal transport correspondence used for human-to-robot retargeting to map interaction targets between demonstrated and substitute objects. Combined with joint pose optimization, this enables our approach to adapt a single demonstration to previously unseen object shapes and sizes, generating multiple retargeted motion variants.

\section{Related Work}\label{sec:rw}

In this section, we review motion retargeting and interaction modeling approaches relevant to whole-body loco-manipulation.

\noindent\textbf{Point-stream retargeting} maps human motion representation, such as joints~\cite{gmr2026}, keypoints~\cite{phc2023}, or a learned latent~\cite{nmr2026}, to robot poses. 
It includes kinematic adaptation to a new skeleton~\cite{gleicher1998}, learned models~\cite{aberman2020,nmr2026} and general-purpose retargeters used in current humanoid pipelines~\cite{gmr2026,phc2023}.
These methods optimize pose tracking and model neither the ground nor manipulated objects; thus, they address none of (i)--(iii).

\noindent\textbf{Relational retargeting} carries the \emph{relative} configuration of body, object and terrain across embodiments.
The original interaction-mesh method tetrahedralizes body joints and the scene vertices per frame, then minimizes the mesh deformation~\cite{interactionmesh2010}. 
Relationship descriptors instead weight sampled surface points by proximity~\cite{alasqhar2013}, interaction meshes were adapted to bipedal locomotion~\cite{imma2012} and manipulation~\cite{toporetarget2026}, and, through a spatial map between two static objects, onto a different object~\cite{kim2016}.
Throughout, the object pose is an input and not a free decision variable~\cite{toporetarget2026}, leaving (i) and (ii) unsolved.
OmniRetarget applies an interaction mesh to whole-body loco-manipulation, but solves only for the robot and globally rescales the demonstration by the robot-to-human height ratio~\cite{omniretarget2025}; a scene-scale extension keeps the same formulation~\cite{meshmimic2026}. Other methods do not model a rigid object at all~\cite{reconform2025,humhumanoid2026}.

\noindent\textbf{Limitations of interaction meshes.} 
Existing relational methods carry interaction through sampled points. In a uniform-weight Laplacian~\cite{omniretarget2025}, each keypoint couples to its mesh neighbors by connectivity~\cite{interactionmesh2010}, not by involvement in the manipulation. Interaction accuracy and posture are coupled in a single metric, and the optimization cost scales poorly with each sampled point. 
Sparse sampling improves tractability at the cost of local contact accuracy. Surface-level relations have been used for self-contact or a static partner in skinned characters~\cite{auramesh2018,villegas2021} but not for a humanoid interacting with a free object.

\noindent\textbf{Dynamics-based recovery} produces dynamically feasible motion through trajectory optimization before tracking \cite{dynaretarget2026}, interaction-aware tracking policies \cite{hdmi2025,resmimic2025}, a bilevel loop around a tracking policy \cite{reactor2026}, or direct consumption of the demonstration~\cite{ddr2026}. They partly address (iii), while (i) and (ii) persist in methods that track a fixed object reference. These approaches focus on the dynamic feasibility of tracking a retargeted motion reference and complement our approach, which aims to preserve a demonstrated interaction. 

\noindent\textbf{Human--object interaction.} We use two ideas from the human--object interaction literature: estimating the object pose jointly with the body~\cite{behave2022} and representing the relation with continuous proximity~\cite{prox2019}.
These methods estimate or synthesize \cite{omomo2023} interactions for an observed object, or imitate them in simulation \cite{intermimic2025}; our goal is to transfer a demonstration to another embodiment. None of the retargeting solvers above jointly decides the object trajectory and carries the demonstrated interaction as a continuous interaction residual.

\section{Method}\label{sec:method}
\subsection{Problem setting and formulation}\label{sec:m-setting}
Conventionally, retargeting pipelines take a human demonstration, typically a video, as input and extract the skeleton and object pose trajectories.
As discussed in Sec.~\ref{sec:intro}, approaches based on human skeleton poses use a coarse approximation of the human geometry, which is insufficient for accurately defining contact.
Our approach instead densely samples points on human and object surfaces to obtain a more detailed geometry representation. 
We similarly sample points on the target robot and objects and \textit{map} them to their corresponding references to specify desired interactions in the target scene.
Deviations from the extracted skeletal and object pose trajectories, which encode the style of motion, and from the target interactions are used as cost terms in a constrained inverse kinematics (IK) problem solved frame by frame:
\begin{equation}
\begin{aligned}
\min_{q_t,\ \{T^{o}_t\}}\quad&
  E^{\mathrm{int}}  + E^{\mathrm{sty}}_{r} + E^{\mathrm{sty}}_o
  + \bigl\lVert q_t \ominus q_{t-1}\bigr\rVert^{2}\\
  \text{s.t.}\quad&
  \mathcal{C}(q_t,\{T^{o}_t\}),
\end{aligned}
\label{eq:m-program}
\end{equation}
where $q_t \in \mathcal{Q} \simeq \mathbb{R}^{n_q}$ is the robot configuration at time $t$, \mbox{$\{T^{o}_t\}\in\mathrm{SE}(3)^{n_{o}}$} the poses of all manipulated objects. $E^{\mathrm{int}}(q_t, \{T^{o}_t\})$ is the deviation cost for the target interaction (Sec.~\ref{sec:m-residuals}), $E^{\mathrm{sty}}_{r}(q_t)$ and $E^{\mathrm{sty}}_o(\{T^{o}_t\})$ are the style deviation costs of the robot and objects, respectively (Sec.~\ref{sec:m-arbitration}).
$\mathcal{C}$~encodes joint limits, joint velocity limits, per-object velocity limits and collision avoidance (including self-collision).

\subsection{Surface-level proximity and optimal transport}\label{sec:m-representation}

To model surface-level interaction between two objects, we define proximity measures for points sampled on the objects' surfaces. Let $\bm{x}_p$, a point in the first object's surface point cloud indexed by $p$, denote a \emph{probe} point. Let the second object, called \emph{channel} $c$, have pose $T^{c}_t = (R_c, \bm{\tau}_c)$.
The probe's position in the channel's frame is $\bm{x}_{c,p}=R_{c}^{\top}(\bm{x}_p-\bm{\tau}_c)$. We define the \textit{proximity triple} for the probe w.r.t.\ the channel as:
\begin{equation}
  \bigl(d,\bm{w},\bm{n}\bigr)_{c,p}
  =\Bigl(\Phi_{c}(\bm{x}_{c,p}),\;W_{c}(\bm{x}_{c,p}),\;
  \tfrac{\bm{x}_{c,p}-\bm{w}_{c,p}}{\lVert
  \bm{x}_{c,p}-\bm{w}_{c,p}\rVert}\Bigr),
  \label{eq:m-triple}
\end{equation}
where $d_{c,p}=\Phi_c(\bm{x}_{c,p})$ is the probe's signed distance to the channel surface, $\bm{w}_{c,p} = W_c(\bm{x}_{c,p})$ the \emph{witness} point, i.e. the closest point on the channel's surface, and $\bm{n}_{c,p}$ the unit vector pointing from witness to probe.
The triple, defined in the channel frame, remains meaningful as the channel moves.
Every entity, i.e. the human, the robot, the objects, carries a probe cloud.
$\Phi_c$ and $W_c$ are precomputed offline on a voxel grid around each object at 1~cm isotropic resolution and up to $\bar{d}=15$~cm from the surface, then linearly interpolated to obtain the value.
The grid is defined in object frame and hence invariant to channel pose transformation and does not require online recomputation. For a flat ground, $\Phi_c$ and $W_c$ are affine.
Probe clouds are sampled at 1000~pts/m$^2$ on objects and 2000~pts/m$^2$ on the human and the robot, and a probe--channel pair is active if the probe--channel distance is less than $\bar{d}$.

\noindent \textbf{Optimal transport.} Retargeting using proximity measures requires correspondences between source and target surfaces.
Each human part is manually pre-assigned to a robot link. For each pair, we cast the correspondence problem between their surface point clouds as an entropically regularized optimal transport problem.
We center and normalize point cloud coordinates by their root-mean-square radius to avoid sensitivity to size differences.
We denote by $a,b$ the uniform marginals derived from the respective objects' point cloud counts and by $C = (C_{ip})$ the matrix of squared Euclidean distances where $i$ and $p$ are the human part and robot link point indices, respectively.
The resulting transport plan is obtained by solving
\begin{equation}
P^{\star}=\arg\min_{P\in \Pi(a,b)}\;\langle P,C\rangle+\varepsilon_{\mathrm{ot}}\textstyle\sum_{i,p}P_{ip}\bigl(\log P_{ip}-1\bigr),
\label{eq:m-sinkhorn}
\end{equation}
where $\Pi(a,b)=\{P\ge 0:\,P\mathbf{1}=a,\;P^{\top}\mathbf{1}=b\}$ is the transport polytope of couplings with marginals $a$ and $b$, $\langle P,C\rangle := \text{tr}(P^\top C)$ denotes the Frobenius inner product, and $\varepsilon_{\mathrm{ot}}=0.1$ is the entropic regularization parameter.
We solve~\eqref{eq:m-sinkhorn} using the Sinkhorn algorithm~\cite{cuturi2013}.
The correspondence for $p$ is assigned as:
\begin{equation}
\mathrm{src}(p)=\arg\max_{i}\,P^{\star}_{ip},
\label{eq:m-ot}
\end{equation}
and the desired robot triples for any channel $c$ follow from the demonstration triples, $(\hat d,\hat{\bm{w}},\hat{\bm{n}})_{c,p}=(d,\bm{w},\bm{n})_{c,\mathrm{src}(p)}$.

Unlike nearest-neighbor assignment, the marginal constraints in problem~\eqref{eq:m-sinkhorn} force every human sample to receive mass, while the entropic term controls how sharply that mass concentrates. Considering top decile points in terms of distance from point cloud centroid, nearest-neighbor assignments leave 42\% of these extremal samples unmatched against 14\% for optimal transport. Coverage is thus more even, in particular over the limb extremities.
The transport is pre-computed once per (human part, robot link) pair by setting both the robot and human in a T-pose, as shown in Fig.~\ref{fig:pipeline}.

\subsection{Interaction residuals}\label{sec:m-residuals}
For any channel $c$ and point $p$, interaction residuals between demonstration and target are defined using the proximity triple:
\begin{equation}
\label{eq:m-residuals}
\begin{aligned}
r^{d}_{c,p} &= d_{c,p}-(\hat d_{c,p})_{+},\\
r^{x}_{c,p} &= \rho_c\!\big(\bm{w}_{c,p}\,;\ \hat{\bm{w}}_{c,p}\big),\\
r^{n}_{c,p} &= \left(\,\operatorname{sign}(\hat d_{c,p})\,\hat{\bm{n}}_{c,p}\!\cdot\!\big(\bm{x}_{c,p}-\hat{\bm{w}}_{c,p}\big)\right)_{-},
\end{aligned}
\end{equation}
where $\rho_c(\bm{w};\hat{\bm{w}})$ is the geodesic distance on the channel surface, $(\cdot)_{+}=\max(\cdot,0)$ and $(\cdot)_{-}=\min(\cdot,0)$.
Geodesic distance is precomputed for each channel surface point cloud, and the distance to an arbitrary witness point is estimated by interpolation.
The demonstrated distance is clamped, $(\hat d_{c,p})_{+}$, so that a noisy demonstration cannot impose a penetration as target. The residual $r^{n}$ penalizes deviation from desired direction of interaction, and resolves the face ambiguities for cases when distance alone does not provide sufficient signal (the wrong side of a thin plate near the edge).
The interaction cost $E^{\mathrm{int}}$ for the IK problem is defined as a weighted sum of quadratic residual losses:
\begin{equation}
\label{eq:m-energy}
\begin{aligned}
E^{\mathrm{int}}=\sum_{c,p}\frac{f_{\ell(p)}}{N_{\ell(p)}}
\Big[&\ {\lambda^{d}}\beta_{c,p}^{2}\,\big(r^{d}_{c,p}\big)^{2}
\\[-7pt]&+\lambda^{x}\bar\beta_{c,p}^{2} \,\big(r^{x}_{c,p}\big)^{2}+{\lambda^{n}}\bar\beta_{c,p}^{2}\big(r^{n}_{c,p}\big)^{2}\Big],
\end{aligned}
\end{equation}
where $N_{\ell(p)}$ is the number of active probes on the link $\ell(p)$ carrying $p$, so that a link's authority does not grow with its sampled area, and $f_{\ell(p)}$ is a per-link weight, kept at $1$ throughout. The weights are $(\lambda^{d},\lambda^{x},\lambda^{n})=(256,625,2500)$ for probes carried by the robot, and four times those for probes carried by an object. The summation is over active channel-probe pairs $(c,p)$ (Sec.~\ref{sec:m-representation}) in either the demonstration or in the retargeted scene: a pair that was close in the demonstration is pulled toward its demonstrated gap; a pair that is close now but was not close in the demonstration is instead pushed back toward that gap.

Both weights $\bar\beta, \beta$ are quadratic kernels of width \mbox{$\sigma=13$\,cm}, so a pair's contribution is attenuated at distances beyond $\sigma$. The first, $\bar\beta=\big(1-(\hat d)_{+}/\sigma\big)_{+}^{2}\le 1$, depends on the demonstration distance $\hat d$ alone, so that $r^{x}$ and $r^{n}$, which determine tangential motion, are active only when the demonstration exhibits contact. The second, \mbox{$\beta=\min\big(\big(1-\min\big((\hat d)_{+},d\big)/\sigma\big)_{+}^{2},\, \beta_{\max}\big)$}, uses the smaller of demonstrated $\hat d$ and the current distance $d$, to control distance when either demonstration or the retarget scene has an active contact pair. Since $d$ is signed, penetration ($d<0$) can drive the kernel above $1$, and is capped at $\beta_{\max}=5$.

\subsection{Joint kinematic retargeting solver}\label{sec:m-arbitration}

\noindent \textbf{Style tracking cost.}
The style cost uses the human skeleton poses as a motion reference through postural costs.
Let $h(\ell)$ map a robot link $\ell$ to the corresponding human link.
For the $L$ tracked robot links, the orientation cost penalizes deviations from the desired orientation reference $\hat R_\ell = R_{h(\ell)}O_\ell$, where the world orientation of the human skeleton $R_{h(\ell)}$ is corrected by a fixed alignment offset $O_\ell$.
We align the robot and human pelvises at $\hat x_0$ with orientation $\hat{R}_\ell$, then compute target link positions by computing forward kinematics over the kinematic tree using the desired orientations and the robot's own link lengths. 
Deriving positions from desired orientations avoids the conflicts between the two that scaling-based methods incur when deforming the human skeleton to the robot's morphology. The link length ratios can range from $0.19$ to $1.18$, with no scaling factor consistent across links. Tracking the full skeleton avoids overweighting the position of any single link. The style costs for links and object frames are
{\setlength{\abovedisplayskip}{3pt}\setlength{\belowdisplayskip}{2pt}%
\begin{align}
E^{\mathrm{sty}}_{r}&=\sum_{\ell}\Bigl[\,{\lambda^{\mathrm{pos}}}\bigl\lVert x_{\ell}(q)-\hat x_{\ell}\bigr\rVert^{2}
+{\lambda^{\mathrm{rot}}}\bigl\lVert R_{\ell}(q)\ominus\hat R_{\ell}\bigr\rVert^{2}\Bigr],
\nonumber\\
E^{\mathrm{sty}}_{o}&=\sum_{o}\Bigl[\,{\lambda^{\mathrm{pos}}_{o}}\bigl\lVert\tau^{o}-\hat\tau^{o}\bigr\rVert^{2}
+{\lambda^{\mathrm{rot}}_{o}}\bigl\lVert R^{o}\ominus\hat R^{o}\bigr\rVert^{2}\Bigr],
\label{eq:m-esty-o}
\end{align}}%
with $R\ominus\hat R={\log(R\hat R^{\top})}^{\vee}$ the world-frame orientation error and $\lambda$ weighting position against orientation.
The style costs carry no robot-object interaction term; the object term centers each manipulated object to its demonstrated trajectory.

\noindent \textbf{Native scale initialization.} The robot's root pose and the scene are retained at the captured metric scale, and the target link positions are computed using forward kinematics. To permit link positions to be determined predominantly by the demonstrated interaction, feet--ground contact included, the position weight $\lambda^{\mathrm{pos}}$ is kept small.
The motion of a human picking an object off a table at a given height must retarget to a robot reaching that same height, regardless of the scale difference; the position targets nevertheless provide a posture prior for links not involved in interaction.

\noindent \textbf{Hard kinematic constraints.} Collision avoidance between the robot, the objects and the ground also uses the signed distance fields as a hard non-penetration constraint $d_{c,p} \ge -\varepsilon_{\mathrm{col}}$ with $\varepsilon_{\mathrm{col}}=0.3$~mm on the sampled points of every probe--channel pair. Self-collision uses capsules approximating the robot links and applies the same criterion.
Each collision pair contributes an inequality constraint in the optimization problem.
In addition to the usual robot joint position and velocity limits, for each object $i$, the boxes $\mathcal{B}_{\mathrm{env}}$ and $\mathcal{B}_{\mathrm{cor}}$ of \eqref{eq:m-qp-box} denote feasible sets that constrain displacement between consecutive frames and deviation from reference motion, respectively: $|\delta\xi_{i} + \xi^{\mathrm{acc}}_{i}| \le b^{\mathrm{env}}_{i}$ and $|\delta\xi_{i} + \xi^{\mathrm{cum}}_{i}| \le b^{\mathrm{cor}}_{i}$, where $\xi^{\mathrm{acc}}_{i}$ is the accumulated single time-step displacement and $\xi^{\mathrm{cum}}_{i}$ the cumulative one since the phase origin.
Fixed objects can be modeled via $b^{\mathrm{cor}}=0$, or, better, by not making their pose a free decision variable in the optimization problem.

\noindent \textbf{Sequential optimization solver.} We solve \eqref{eq:m-program} using a trust-region sequential quadratic programming (SQP) approach. At each SQP iterate, we employ a  Gauss--Newton Hessian approximation and linearize all constraints to obtain a quadratic program (QP) as the inner problem.  The QP's decision variables are step changes between SQP iterations in robot configuration $\delta v$ (floating base and joints) and object pose $\delta\xi_o=(\delta\tau_o,\delta\theta_o)\in\mathbb{R}^6$ per manipulated object, retracted as $\tau^o\leftarrow\tau^o+\delta\tau_o$, $R^o\leftarrow\exp(\delta\theta_o)\,R^o$. Every manipulated object is a decision variable.
 The QP subproblem is:
\begin{subequations}\label{eq:m-qp}
{\setlength{\abovedisplayskip}{3pt}\setlength{\belowdisplayskip}{3pt}%
\begin{align}
\min_{\delta v,\,\delta\xi}\quad & \widehat{E}^{\mathrm{sty}}_{r}+\widehat{E}^{\mathrm{sty}}_{o}+\widehat{E}^{\mathrm{int}}
  + \bigl\lVert (q_{t} \oplus \delta v) \ominus q_{t-1} \bigr\rVert^{2}
  \label{eq:m-qp-obj}\\
\text{s.t.}\quad
& q^{\min}\le q_{t}\oplus\delta v\le q^{\max},\label{eq:m-qp-lim}\\
& \bigl|(q_{J} \oplus S\delta v) \ominus q_{J,t-1}\bigr|\le v_{\max}\,\Delta t,\label{eq:m-qp-vel}\\
& d_{c,p}+\bm{g}_{c,p}^{\top}(\delta v,\delta\xi)\ \ge\ -\varepsilon_{\mathrm{col}},
  \label{eq:m-qp-col}\\
& \delta\xi\in\mathcal{B}_{\mathrm{env}}\cap\mathcal{B}_{\mathrm{cor}},\label{eq:m-qp-box}\\
& \lvert\delta v\rvert\le\Delta_{v},\ \ \lvert\delta\xi\rvert\le\Delta_{\xi},\label{eq:m-qp-tr}
\end{align}}%
\end{subequations}
where $\widehat{E}^{\mathrm{sty}}_{r},\widehat{E}^{\mathrm{sty}}_{o},\widehat{E}^{\mathrm{int}}$ are the Gauss--Newton quadratic objective approximations, $q_{J}$ the actuated joint angles at the iterate, $S$ their selector in $\delta v$, $v_{\max}$ the joint velocity limits, and $\Delta_v,\Delta_\xi$ fixed per-DoF boxes ($5$~cm, $0.10$~rad).
The loop runs a fixed budget of $50$ iterations on the cold-started first frame and six warm-started thereafter, or stops when the step falls below $10^{-4}$; the QP is assembled from the analytic frame Jacobians of Pinocchio~\cite{carpentier2019pinocchio}
and solved with ProxQP~\cite{proxqp2022}, warm-started across SQP iterations and frames.
\subsection{Generalization to new object shapes}\label{sec:m-datagen}
Our approach extends to new object shapes and sizes using a single demonstration. We hypothesize that preserving the approach motion and contact relationships enables interaction transfer between objects with similar surface topology. To this end, we reuse the optimal transport formulation introduced for human-to-robot correspondence in Sec.~\ref{sec:m-representation} to map interaction targets from the demonstrated object to a substitute object.

We precompute correspondences between the two surface point clouds, each expressed in its object's local frame and normalized to unit radius.
This normalization is used only to establish correspondences: witness locations are mapped onto the substitute surface, while the interaction targets retain their metric meaning without rescaling.

The mapped targets are then incorporated into the joint retargeting problem, with the substitute object's pose as a decision variable. This allows robot--object and object--object relative motions to adapt to the new geometry while preserving the demonstrated interactions.

\section{Experiments}\label{sec:exp}

We evaluate \sys{} on motion and interaction fidelity, runtime, object generalization, and hardware transfer through learned tracking policies.

\subsection{Experimental setup and evaluation protocol}\label{sec:e-protocol}

\noindent\textbf{Robot, demonstrations, and baselines.}
All demonstrations are retargeted to the 29-DoF Unitree G1 with fixed hands, for which a \emph{grip} is defined by hand--object proximity.
Our evaluation covers all $2\,027$ body-only motions from the CMU and SFU subsets of AMASS~\cite{amass2019}, all $4\,421$ single-object manipulation sequences from 13 OMOMO categories~\cite{omomo2023} in the InterMimic release~\cite{intermimic2025}, and our own two-object RGB-D sequence (\emph{pnp14}).
For the latter, we recover human motion with GVHMR~\cite{shen2024gvhmr} and object poses with FoundationPose~\cite{wen2024foundationpose}, using primitive meshes at measured dimensions for object geometry.
We compare against OmniRetarget~\cite{omniretarget2025}, GMR~\cite{gmr2026}, and PHC~\cite{phc2023} under their released settings, marking any adaptations. Since GMR and PHC do not model objects, they are excluded from OMOMO.
OmniRetarget reports infeasibility on $424$ OMOMO sequences, mostly involving chairs and large tables. Omitting these failures from its aggregate metrics slightly favors the baseline.

Each method retains its scaling convention: OmniRetarget and our scaled ablation scale human and object trajectories by $sc=\text{robot height}/\text{stature}$ without resizing objects, where G1 is $1.32$\,m tall and stature denotes the \mbox{SMPL-X} rest-mesh height. GMR uses its released root-path scale, $sc=0.9\times\text{stature}/1.8$, while PHC and our native-scale variants use $sc=1$. We report the median $sc$ for each dataset.

\noindent\textbf{Interaction metrics.}
We represent each interaction by a sample $(t,u,c)$ pairing a body unit $u$ with a channel $c$ (ground or object) at frame $t$, and record its signed distance and closest surface point in the demonstration, $(\hat d,\hat{\bm w})$, and on the robot, $(d,\bm w)$.
For all methods, these quantities are computed independently of the solver using exact signed distances from posed \mbox{SMPL-X} and robot visual-mesh vertices to the channel mesh, with each unit's distance determined by its closest vertex.

With a threshold of $\varepsilon_{\mathrm{eval}}=20$\,mm, the demonstrated and retargeted interaction sets are \mbox{$\mathcal H=\{(t,u,c)\,|\,(\hat d)_+<\varepsilon_{\mathrm{eval}}\}$} and
$\mathcal R=\{(t,u,c)\,|\,(d)_+<\varepsilon_{\mathrm{eval}}\}$.
We measure their agreement through precision
$P=|\mathcal{H}\cap\mathcal{R}|/|\mathcal{R}|$, recall
$R=|\mathcal{H}\cap\mathcal{R}|/|\mathcal{H}|$ and Jaccard index
$J=|\mathcal{H}\cap\mathcal{R}|/|\mathcal{H}\cup\mathcal{R}|$. 
To assess geometric accuracy over $\mathcal H$, we also report depth error
$dd=\mathrm{mean}\,|d-(\hat d)_+|$ and placement error
$dw=\mathrm{mean}\,\lVert\bm w-\hat{\bm w}\rVert$, measuring deviations in gap and surface location, respectively. Ground and object interactions are reported separately; penetrations count as interactions, with their magnitude captured by $dd$ on $\mathcal H$.

\noindent\textbf{Additional metrics.}
We assess motion style using \emph{rot}, the mean link-orientation error in degrees relative to the style target, with the same SMPL-X-to-robot alignment $O_\ell$ for all methods. Position error is omitted because leg-length differences dominate it.
For self-collision, \emph{self} measures the 95th-percentile penetration depth from the robot visual mesh, independently of the solver's capsules.
Foot contact is characterized by \emph{skate}, the median sliding velocity of a foot's least-moving point during stance, and \emph{clear}, its median stance height.
Finally, \emph{drift} measures mean object-position deviation from the demonstration and is zero for fixed trajectories.
Our objective does not directly optimize \emph{self}, \emph{skate}, or \emph{clear}.

\noindent\textbf{Aggregation and parameter settings.}
Within each sequence, we compute $P$, $R$, and $J$ as sample ratios, average $dd$ and $dw$ over $\mathcal H$, and average \emph{rot} and \emph{drift} over frames. We then report dataset medians over the sequences returned by each method.
\sys{} uses the same weights across datasets and scenes, without per-sequence tuning. Baselines retain their released settings, except for the object-interaction weights in our OmniRetarget extension.

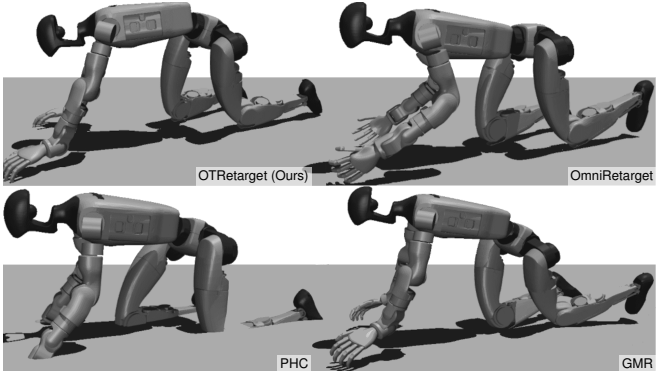
\begin{figure}[t]
\vspace{1.0em}
\centering
\begin{tikzpicture}
  \node[anchor=south west,inner sep=0] (img)
    {\includegraphics[width=\linewidth, trim={15pt 2pt 0pt 7pt}, clip]{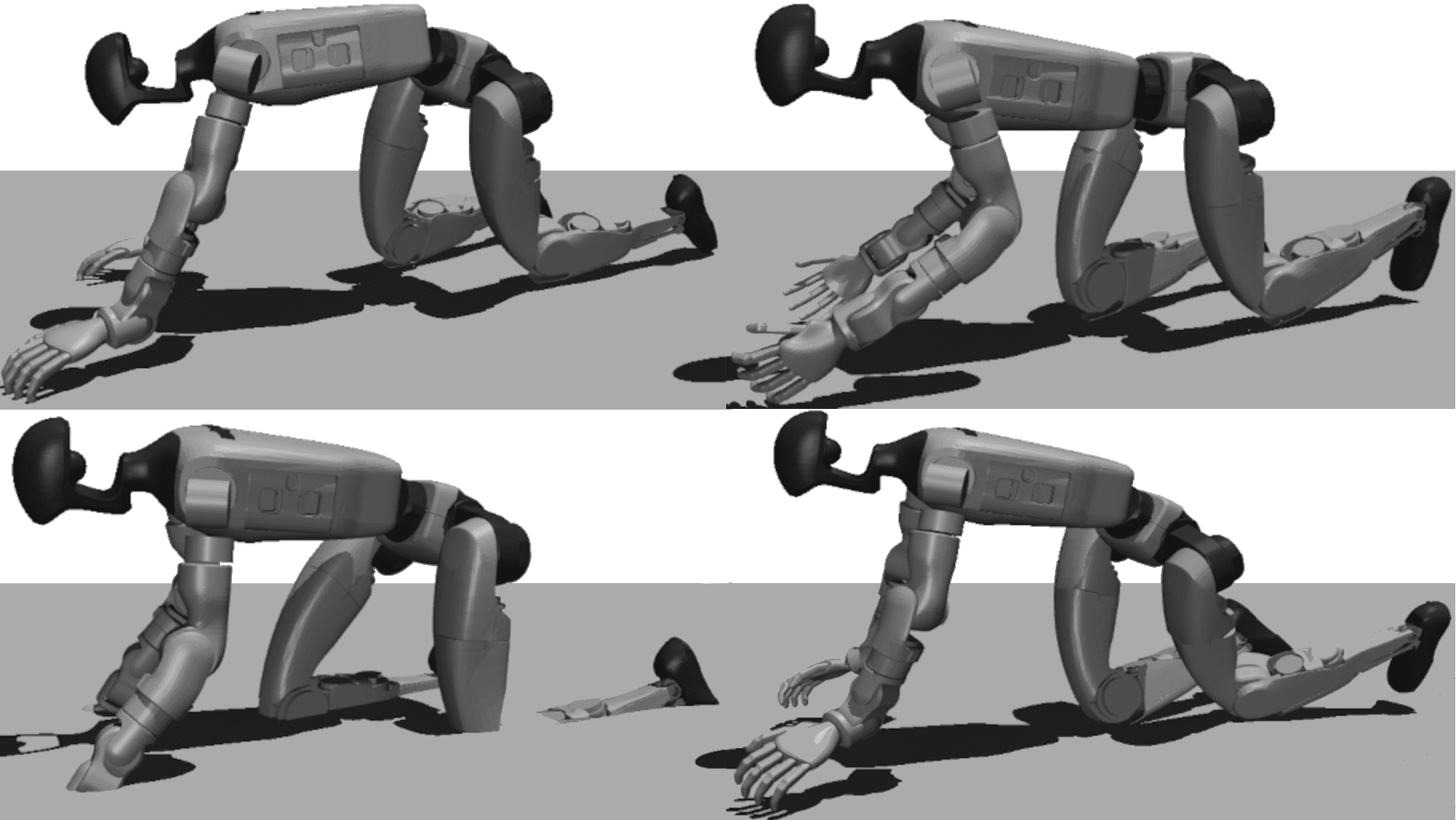}};
  \begin{scope}[x={(img.south east)},y={(img.north west)},
                every node/.style={font=\tiny\sffamily, inner sep=1pt,
                                   fill=white, fill opacity=0.60, text opacity=1,
                                   anchor=north west}]
    \node at (0.295,0.552) {OTRetarget (Ours)};
    \node at (0.864,0.552) {OmniRetarget};
    \node at (0.42,0.044) {PHC};
    \node at (0.943,0.044) {GMR};
  \end{scope}
\end{tikzpicture}
\caption{\textbf{Hands and knees on the floor} (\emph{crawl}). \sys{} lays both hands flat
on the floor as the demonstration does. OmniRetarget reaches the floor with the wrists
bent; GMR, which represents no terrain, leaves the hands above it, and PHC drives them
through it.}
\label{fig:crawl}
\vspace{-1em}
\end{figure}

\subsection{Comparison with existing retargeting methods}\label{sec:e-sota}

\begin{figure}[t]
\vspace{1.0em}
\centering
\begin{tikzpicture}
  \node[anchor=south west,inner sep=0] (img)
    {\includegraphics[width=\linewidth]{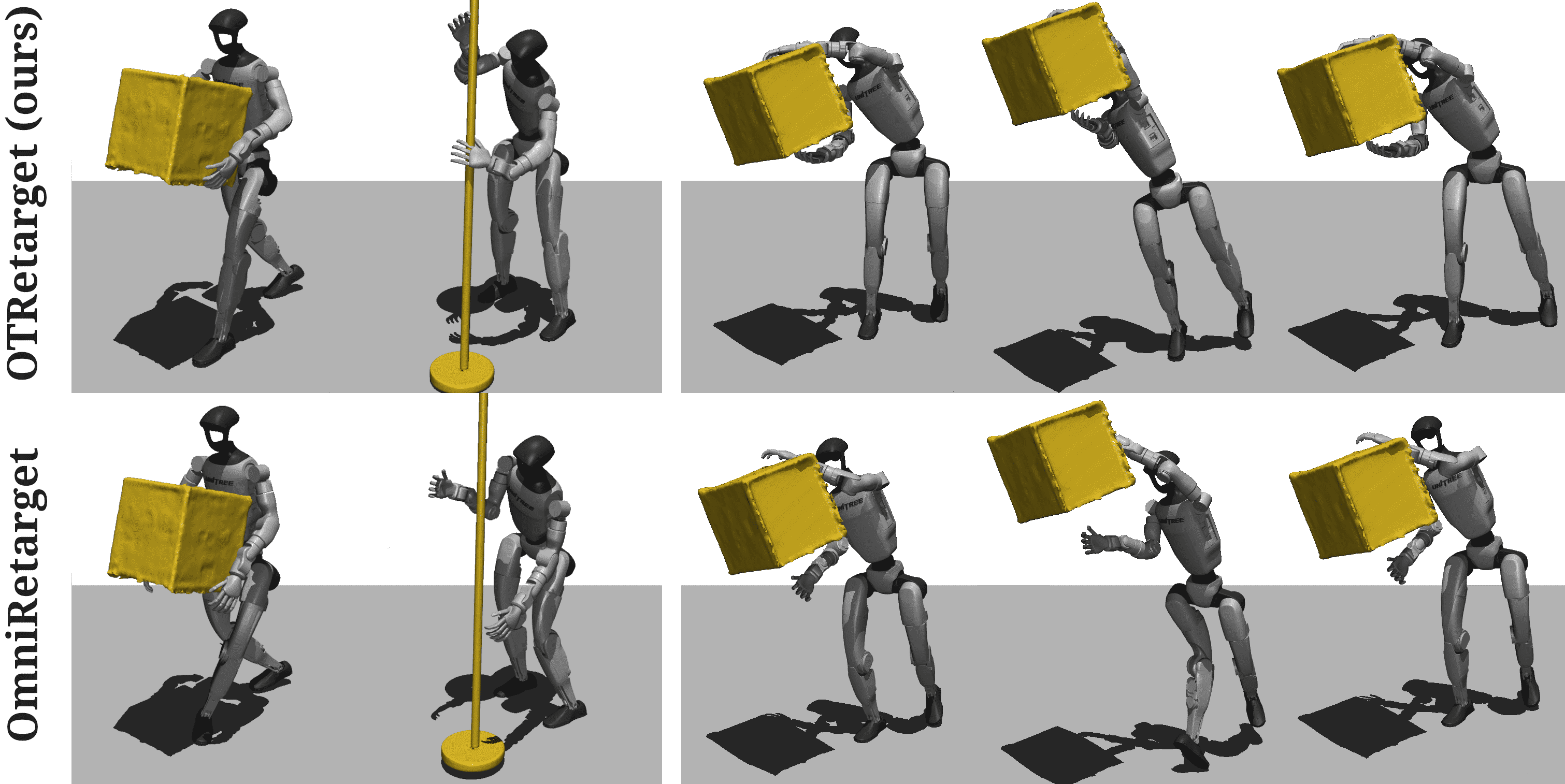}};
  \begin{scope}[x={(img.south east)},y={(img.north west)},
                every node/.style={font=\fontsize{4}{5}\selectfont\sffamily,
                                   inner sep=0.5pt, anchor=north,
                                   fill=white, fill opacity=0.7, text opacity=1}]
    \node at (0.545,0.538) {Scaled, obj. fixed};
    \node at (0.715,0.538) {Obj. fixed};
    \node at (0.900,0.538) {Obj. variable};
    \node at (0.545,0.041) {Scaled, obj. fixed};
    \node at (0.715,0.041) {Obj. fixed};
    \node at (0.900,0.041) {Obj. variable};
  \end{scope}
\end{tikzpicture}
\caption{\textbf{Illustration of loco-manipulation on OMOMO,} \sys{} (top) against OmniRetarget (bottom): a large convex box, then a floor lamp, a stem the mesh has no volume to hold (left); the same overhead lift in the scaled world, native with the object pinned, and native with it free (right).}
\label{fig:manip-panel}
\vspace{-1em}
\end{figure}
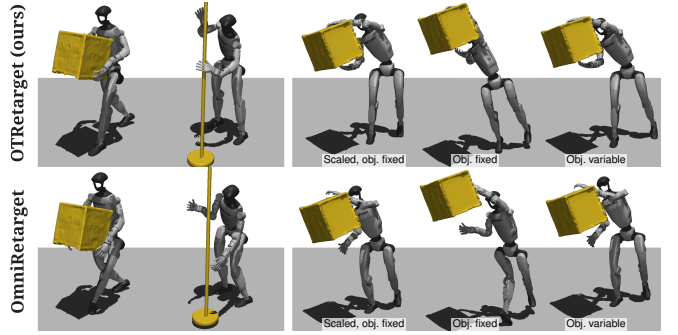

\noindent\textbf{Ground interaction.}
\cref{tab:scaled-locomotion} evaluates every method over $2\,027$ AMASS and $4\,421$ OMOMO sequences. 
\sys{} achieves the best style error, floor depth, and interaction agreement on both datasets (\cref{fig:crawl}).
The \emph{rot} gap to OmniRetarget and PHC is structural rather than a matter of tuning: both optimize joint positions without directly constraining link orientation, which causes the wrists to remain bent on the floor (\cref{fig:crawl}) and the feet splayed outward during carrying (\cref{fig:manip-panel}). 
GMR has the lowest \emph{skate}, but its soles never settle on the floor (\emph{clear}); OmniRetarget has the opposite failure with negative clearance.
On OMOMO, the two interaction-aware methods retain ground contact while their style error increases because the robot must adapt its grip. OmniRetarget reaches $13.7$\,mm self-penetration on AMASS because self-collision is disabled by default; GMR and PHC constrain neither self- nor scene-collision.

\begin{table}[b]
\vspace{-1em}
\centering
    \captionsetup{labelsep=colon}
\caption{\textbf{Ground fidelity} on locomotion (AMASS, CMU and SFU pooled) and loco-manipulation (OMOMO), each method in its released world; dataset medians, sequence counts in the block headers.}
\label{tab:scaled-locomotion}
\footnotesize
\setlength{\tabcolsep}{2.2pt}
\begin{tabular}{lrrrrrrrrr}
\toprule
& & & & \multicolumn{6}{c}{Terrain} \\
\cmidrule(lr){5-10}
Method & $sc$ & rot$\downarrow$ & self$\downarrow$ & dd$\downarrow$ & R$\uparrow$ & P$\uparrow$ & J$\uparrow$ & skate$\downarrow$ & clear \\
& & $^\circ$ & mm & mm & \% & \% & \% & cm/s & mm \\
\midrule
\multicolumn{10}{l}{\textit{Locomotion - AMASS (2027 sequences)}} \\
\midrule
 \sys{} & 1.00 & \textbf{6.3} & \textbf{0.0} & \textbf{1.2} & \textbf{99} & 100 & \textbf{99} & 1.8 & 1.3 \\
 OmniRetarget & 0.76 & 25.5 & 13.7 & 3.3 & 95 & 100 & 93 & 2.1 & 5.1 \\
 GMR & 0.87 & 7.5 & \textbf{0.0} & 23.2 & 47 & 100 & 47 & \textbf{1.3} & 30.9 \\
 PHC & 1.00 & 21.7 & \textbf{0.0} & 17.3 & 68 & 100 & 68 & 4.9 & 29.7 \\
\midrule
\multicolumn{10}{l}{\textit{One-Object Manipulation - OMOMO (4421 sequences)}} \\
\midrule
 \sys{} & 1.00 & \textbf{10.7} & \textbf{0.0} & \textbf{1.7} & \textbf{99} & 100 & \textbf{99} & 3.5 & 4.6 \\
 OmniRetarget & 0.75 & 27.8 & 2.0 & 2.7 & 98 & 100 & 96 & \textbf{3.0} & -0.7 \\
\bottomrule
\end{tabular}
\end{table}

\noindent\textbf{Robot-object interaction.}
\cref{tab:scaled-manipulation} reports robot--object interaction on OMOMO.
\sys{} reaches $87\%$ Jaccard agreement, versus $28\%$ for OmniRetarget, with one third of its depth and placement errors.
The scaled, fixed-object ablation retains most of this gain, showing that the interaction residuals, rather than the native scene or variable object, drive the improvement. 

On a large convex box, OmniRetarget matches the demonstrated depth more closely but also predicts contact in non-contact frames, lowering precision (\cref{fig:manip-panel}, left). On a floor lamp, whose stem affords a small grasp volume, it loses the interaction entirely; our result is unchanged. Thus, OmniRetarget's interaction mesh appears to be more sensitive to object geometry than \sys{}'s surface stream.

\noindent\textbf{Runtime.}
Our unoptimized Python implementation achieves median runtimes of $21.5$\,ms per frame for locomotion (\emph{spin, CMU, 88\_10}) and $63.8$\,ms for object carrying (\emph{largebox, OMOMO, sub3\_003}), measured on a single core of an AMD Threadripper PRO 7955WX.
On these sequences, our approach is $12$--$22\times$ faster than OmniRetarget, which requires $470$ and $749$\,ms per frame, respectively. GMR remains faster at $3.2$ and $2.5$\,ms, but does not model interactions with the environment. These results motivate further implementation optimization toward online retargeting.

\subsection{Retargeting the native scene}\label{sec:e-native}

\begin{table}[b]
\vspace{-1em}
\centering
    \captionsetup{labelsep=colon}
\caption{\textbf{The object as a decision variable.} The scene solved as captured (ablation: object fixed vs.\ variable); \emph{drift} is blank on scaled rows, where it would measure the world's scale, not the method.}
\label{tab:ablation}
\footnotesize
\setlength{\tabcolsep}{2.2pt}
\begin{tabular}{lrrrrrrr}
\toprule
 & & & Terrain & \multicolumn{4}{c}{Object} \\
\cmidrule(lr){4-4}\cmidrule(lr){5-8}
Method & $sc$ & rot$\downarrow$ & R$\uparrow$ & R$\uparrow$ & P$\uparrow$ & J$\uparrow$ & drift \\
 & & $^\circ$ & \% & \% & \% & \% & cm \\
\midrule
\multicolumn{8}{l}{\textit{OMOMO (4421 sequences)}} \\
\midrule
\sys{} (scaled, obj.\ fixed) & 0.75 & 11.1 & \textbf{100} & \textbf{97} & 89 & 82 & --- \\
\hspace{4pt}native scene, obj.\ fixed & 1.00 & 12.0 & 98 & 94 & 96 & 84 & 0.0 \\
\hspace{4pt}native scene, obj.\ variable & 1.00 & \textbf{10.7} & 99 & 95 & 96 & \textbf{87} & 2.9 \\
\midrule
\multicolumn{8}{l}{\textit{largebox, overhead lift (OMOMO, sub8\_028)}} \\
\midrule
\sys{} (scaled, obj.\ fixed) & 0.72 & \textbf{8.8} & \textbf{100} & 72 & 99 & 72 & --- \\
\hspace{4pt}native scene, obj.\ fixed & 1.00 & 16.8 & 79 & 71 & 98 & 71 & 0.0 \\
\hspace{4pt}native scene, obj.\ variable & 1.00 & 9.8 & 99 & \textbf{80} & 99 & \textbf{79} & 12.2 \\[3pt]
OmniRetarget & 0.72 & 27.0 & 99 & 48 & 82 & 44 & --- \\
\hspace{4pt}native scene (our ext.) & 1.00 & 34.7 & 48 & 46 & 79 & 41 & 0.0 \\
\hspace{4pt}+ ground in the graph (our ext.) & 1.00 & 48.8 & 74 & 46 & 62 & 36 & 0.0 \\
\hspace{4pt}+ obj.\ variable (our ext.) & 1.00 & 28.8 & \textbf{100} & 54 & 71 & 44 & 32.4 \\
\bottomrule
\end{tabular}
\end{table}

\cref{tab:ablation} reports results for retargeting in a captured, native-scale scene and the benefits of a variable object. On OMOMO, fixing the object slightly worsens style and ground contact; freeing it recovers both with less than $3$\,cm drift. 
The median hides the important overhead-lift case (\cref{fig:manip-panel}, right): a fixed native object doubles the style error and lifts the feet, whereas a variable object moves $12$\,cm on average to a reachable height and recovers posture, stance, and grip. 
Our OmniRetarget extension with a variable object shows a similar trend. Adding ground to its graph improves ground contact but worsens posture; freeing the object improves both at the cost of almost three times our drift. But object grip remains poor, indicating the importance of the interaction term.

\noindent\textbf{Multi-object retargeting.}
In \emph{pnp14} (\cref{fig:teaser}a and \cref{fig:pipeline}) a box is lifted from the floor onto a table; the two objects must make appropriate contact when the box comes to rest.
Box-table and box-arm interactions' recall and precision remain high at $95\%$ and $98\%$. The table object remains on the floor and the robot does not make spurious table contact. In flight, the box departs from its demonstrated trajectory by up to $27$\,cm to remain reachable, then settles on the table at the demonstrated place during demonstrated frames, with a $2.7$\,cm mean drift and an $8.8^\circ$ \emph{rot} error.

\begin{table}[t]
\centering
\vspace{0.5em}

    \captionsetup{labelsep=colon}
\caption{\textbf{Robot--object proximity} on the OMOMO dataset (medians) and on two single captures, a large convex box and a floor lamp; the marked row solves in OmniRetarget's scaled world with the object fixed.}
\label{tab:scaled-manipulation}
\footnotesize
\setlength{\tabcolsep}{1.7pt}
\begin{tabular}{lrrrrrr}
\toprule
& & \multicolumn{5}{c}{Object} \\
\cmidrule(lr){3-7}
Method & $sc$ & dd$\downarrow$ & dw$\downarrow$ & R$\uparrow$ & P$\uparrow$ & J$\uparrow$ \\
 & & mm & mm & \% & \% & \% \\
\midrule
\multicolumn{7}{l}{\textit{OMOMO (4421 sequences)}} \\
\midrule
\sys{} & 1.00 & \textbf{8.7} & \textbf{37} & 95 & 96 & \textbf{87} \\
\sys{} (scaled, obj.\ fixed) & 0.75 & 8.8 & 38 & \textbf{97} & 89 & 82 \\
OmniRetarget & 0.75 & 29.3 & 102 & 39 & 62 & 28 \\
\midrule
\multicolumn{7}{l}{\textit{largebox (OMOMO, sub3\_003)}} \\
\midrule
\sys{} & 1.00 & 10.0 & \textbf{42} & 94 & 89 & \textbf{85} \\
\sys{} (scaled, obj.\ fixed) & 0.68 & 9.1 & 44 & \textbf{97} & 81 & 79 \\
OmniRetarget & 0.68 & \textbf{5.7} & 54 & 95 & 58 & 56 \\
\midrule
\multicolumn{7}{l}{\textit{floorlamp (OMOMO, sub10\_031)}} \\
\midrule
\sys{} & 1.00 & \textbf{9.3} & \textbf{37} & \textbf{99} & 100 & \textbf{99} \\
\sys{} (scaled, obj.\ fixed) & 0.80 & 9.4 & 38 & \textbf{99} & 100 & \textbf{99} \\
OmniRetarget & 0.80 & 108.6 & 302 & 0 & --- & 0 \\
\bottomrule
\end{tabular}
\vspace{-1em}
\end{table}

\subsection{Retargeting generalization to object variations}\label{sec:e-datagen}

\begin{figure}[b]
\vspace{-1em}
\centering
\includegraphics[width=\linewidth, trim={6pt 4pt 2pt 10pt}, clip]{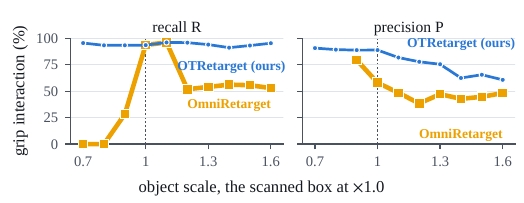}
\vspace{-1em}
    \captionsetup{labelsep=colon}
\caption{\textbf{Illustration of one demonstration replayed across a distribution of object sizes} (\emph{largebox, OMOMO, sub3\_003}): grip recall and precision against the scale of the box.}
\label{fig:sweep}
\end{figure}

\noindent\textbf{Object resizing.}
\cref{fig:sweep} resizes the scanned box from $\times0.7$ to $\times1.6$.
 \sys{}'s grip recall remains nearly constant because the proximity triples evaluated on the substituted surface and variable object enable contact at desired time instances. 
OmniRetarget is accurate only at its native size: recall collapses on smaller boxes and halves with OmniRetarget's Laplacian metric not sufficiently encoding desired contact as its mesh vertices are scaled.
Our precision decreases for larger boxes because contact is made for a longer period, induced by style targets.

\noindent\textbf{Object swapping.}
\cref{tab:datagen} replaces the scanned box with eight unseen shapes, from a ball to a beam.
The demonstration remains unchanged, while the object$\leftrightarrow$object transport plan of Sec.~\ref{sec:m-datagen} maps each demonstrated witness to the substitute.
Recall and precision remain near those of the native box and the depth error stays within $3$\,mm. Drift increases for shapes less similar to a box, reaching roughly twice the native value for the torus and board. This drift is the adaptation required to preserve the interaction, not an error that a fixed trajectory could remove.

\subsection{Policy training and hardware transfer}\label{sec:e-policy}

\noindent\textbf{Policy training setup.}
The reinforcement learning policy is used only for downstream evaluation of reference quality and does not constitute a contribution.
We use the Holosoma framework~\cite{omniretarget2025} and its published PPO~\cite{ppo2017} recipe, which uses $4096$ environments, $30\,000$ iterations. 
An episode succeeds when it completes the clip without a tracking termination (a tracked body/torso deviates from its reference by more than $25/50$~cm). Evaluations use the mean action.

\begin{figure}[t]
\centering
\includegraphics[width=\linewidth,trim={6pt 7pt 0pt 2pt}, clip]{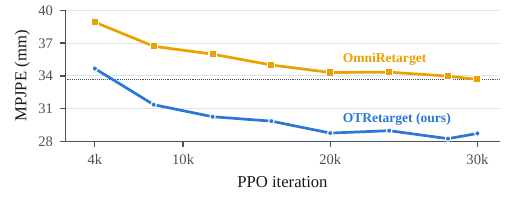}
\caption{\textbf{Impact of Retargeting Quality on Tracking Performance.} MPJPE against PPO iterations for ten AMASS clips using identical training pipelines. The dotted line marks OmniRetarget's final accuracy.}
\label{fig:policy}
\vspace{-1em}
\end{figure}

\noindent\textbf{Locomotion.}
Without an object, the original framework is used. We retarget ten AMASS clips, from crawling to spinning on one leg, with both methods and train a policy per clip under identical settings. Success rates are similar ($98.8\%$ vs.\ $97.2\%$), but tracking error differs significantly. Rolling out every saved checkpoint against its own reference, we measure mean per-joint position error
\mbox{$\mathrm{MPJPE}=\tfrac{1}{|\mathcal{B}|}\sum_{b\in\mathcal{B}}\bigl\|(\bm{p}_b-\bm{p}_{\mathrm{root}})-(\hat{\bm{p}}_b-\hat{\bm{p}}_{\mathrm{root}})\bigr\|$}
over the $|\mathcal{B}|=14$ tracked bodies, simulated $\bm{p}$ against reference $\hat{\bm{p}}$, each relative to its own root so it reads posture, not path.
Policies trained on our reference reach $28.8$~mm versus $33.7$~mm for OmniRetarget and lead by $\approx 5$~mm throughout training (Fig.~\ref{fig:policy}). The improvement is not due to an easier reference, as ours is closer to the demonstration (Tab.~\ref{tab:scaled-locomotion}). Using 
OmniRetarget's final MPJPE as a threshold, our policies reach it by iteration $6$k, at one fifth of the baseline budget.

\begin{table}[t]
\centering
\vspace{0.5em}

    \captionsetup{labelsep=colon}
\caption{\textbf{One capture, multiple unseen objects:} the capture never touched (\emph{largebox, OMOMO, sub3\_003}), every row is obtained with \sys{}}
\label{tab:datagen}
\footnotesize
\setlength{\tabcolsep}{2.2pt}
\begin{tabular}{lrrrrr}
\toprule
 & \multicolumn{4}{c}{Object} & \\
\cmidrule(lr){2-5}
Variant & dd$\downarrow$ & R$\uparrow$ & P$\uparrow$ & J$\uparrow$ & drift \\
 & mm & \% & \% & \% & cm \\
\midrule
box $\times$1.0 (native) & 10.0 & 94 & 89 & 85 & 9.9 \\
\midrule
ball $\varnothing$0.34 & 12.1 & 97 & 87 & 84 & 12.4 \\
drum $\varnothing$0.34$\times$0.36 & 12.2 & 95 & 88 & 84 & 11.1 \\
capsule $\varnothing$0.23$\times$0.36 & 11.8 & 96 & 88 & 84 & 13.1 \\
rugby ball 0.43$\times$0.31$\times$0.28 & 9.4 & 97 & 86 & 84 & 13.8 \\
torus $\varnothing$0.36 & 10.2 & 97 & 90 & 87 & 18.6 \\
cube 0.26 & 9.9 & 89 & 90 & 81 & 13.3 \\
board 0.38$\times$0.30$\times$0.12 & 10.1 & 96 & 88 & 85 & 16.8 \\
beam 0.46$\times$0.19$\times$0.19 & 12.0 & 94 & 87 & 82 & 14.7 \\
\bottomrule
\end{tabular}
\vspace{-1em}
\end{table}

\noindent\textbf{Manipulation.}
For manipulation, we demonstrate transfer rather than compare references. 
Holosoma's released object framework plateaus at $2\%$ success on \emph{pnp14} with our native, variable-object reference. Training requires object reference-aware addition to Holosoma, making an A/B comparison with the Holosoma framework misleading. We add box linear- and angular-velocity tracking and the force-set point reward of~\cite{hdmi2025}, gated by robot--box and box--table proximity windows from the reference. Training reaches $48\%$ success under a $25$~cm object termination, with the box $11$~cm from its reference. The same policy runs on the physical G1 without retraining or object perception and completes $6$ of $8$ pick-and-place trials (supplementary video). This is a single-robot, single-clip demonstration rather than a hardware study.

\section{Conclusion}\label{sec:conclusion}
We have introduced \sys{}, a geometry-aware approach to jointly retarget human and multi-object motion to humanoid robots.
Our formulation combines surface-level interaction targets with optimal transport to adapt robot and object trajectories while preserving contacts without rescaling the scene.
Across two datasets comprising $6\,448$ clips, our approach improves both motion-style tracking and interaction fidelity over existing methods and supports transfer to unseen object shapes and sizes.
Whole-body policies trained on the resulting references transfer to a physical G1 humanoid, including for two-handed box pick-and-place.
Our evaluation remains limited to one robot, flat ground, and a single two-object capture. The formulation is kinematic and operates frame by frame, without explicitly modeling forces or optimizing over a time horizon.
A promising direction could be adequately accounting for these modalities to embrace more complex terrains and scenarios.

\section*{Acknowledgments}
This work was supported by the European Union's Horizon Europe research and innovation programme through the Marie Sk\l{}odowska-Curie Postdoctoral Fellowship ExTRAORDiNary (grant no.\ 101211945) and the ARTIFACT project (grant no.\ 101165695). Additional support was provided by the French government through the ``PR[AI]RIE-PSAI'' AI Cluster (ANR-23-IACL-0008), managed by the Agence Nationale de la Recherche, and through the France 2030 Organic Robotics Program (PEPR O2R) and the PIQ program, the latter managed by the Agence de Programme du Num\'erique.
Views and opinions expressed are those of the author(s) only and do not necessarily reflect those of the European Union or the European Commission. Neither the European Union nor the European Commission can be held responsible for them.

{
\balance
\bibliographystyle{IEEEtran}
\bibliography{references}
}

\end{document}